\pdfoutput=1

\documentclass[11pt]{article}

\usepackage{emnlp2021}

\usepackage{times}
\usepackage{latexsym}

\usepackage[T1]{fontenc}

\usepackage[utf8]{inputenc}

\usepackage{microtype}

\usepackage{graphicx} 
\usepackage{float} 
\usepackage{subfigure} 
\usepackage{url}
\usepackage{amsmath}
\usepackage{algorithm}
\usepackage{algorithmicx}
\usepackage{algpseudocode}
\usepackage{color}
\usepackage{booktabs}
\usepackage{multirow}
\floatname{algorithm}{Algorithm}

\title{Enhanced Language Representation with Label Knowledge for \\ Span Extraction}

\author{Pan Yang$^{1,2}$\thanks{\hspace{0.15cm}{This work was done in ICT, CAS.}}, \  Xin Cong$^{3,4}$, \  Zhenyun Sun$^{1,5}$, \ \bf{Xingwu Liu}${^6}$\thanks{\hspace{0.15cm}Corresponding Author}  \\
	$^1$Institute of Computing Technology, Chinese Academy of Sciences \\
	$^2$PCG, Tencent \\
	$^3$Institute of Information Engineering, Chinese Academy of Sciences \\
	$^4$School of Cyber Security, University of Chinese Academy of Sciences \\
	$^5$School of Computer Science and Technology, University of Chinese Academy of Sciences \\
	$^6$School of Mathematical Sciences, Dalian University of Technology\\
	{\tt im.panyang@gmail.com}, {\tt congxin@iie.ac.cn} \\
	{\tt sunzhenyu@ict.ac.cn}, 	{\tt liuxingwu@dlut.edu.cn}
}

\begin{document}
\maketitle
\begin{abstract}

Span extraction, aiming to extract text spans (such as words or phrases) from plain texts, is a fundamental process in Information Extraction. Recent works introduce the label knowledge to enhance the text representation by formalizing the span extraction task into a question answering problem (QA Formalization), which achieves state-of-the-art performance.
However, QA Formalization does not fully exploit the label knowledge and suffers from low efficiency in 
 training/inference.
To address those problems, we introduce a new paradigm to integrate label knowledge and further propose a novel model to explicitly and efficiently integrate label knowledge into text representations.
Specifically, it encodes texts and label annotations independently and then integrates label knowledge into text representation with an elaborate-designed semantics fusion module. 
%
%
We conduct extensive experiments on three typical span extraction tasks: flat NER, nested NER, and event detection. The empirical results show that 1) our method achieves state-of-the-art performance on four benchmarks, and 2) reduces training time and inference time by 76\% and 77\% on average, respectively, compared with the QA Formalization paradigm. Our code and data are available at \url{https://github.com/Akeepers/LEAR}.
\end{abstract}

\section{Introduction}
Information Extraction~(IE), a fundamental task in natural language processing, aims to extract structured knowledge from  unstructured texts. It usually contains the process that extracts text spans (such as words or phrases) from plain text, e.g., NER. Span extraction is usually formulated into the sequence labeling problem that assigns a categorical label to each token in a text. 

\begin{figure}[!t]
	\centering
	\includegraphics[width=0.40\textwidth]{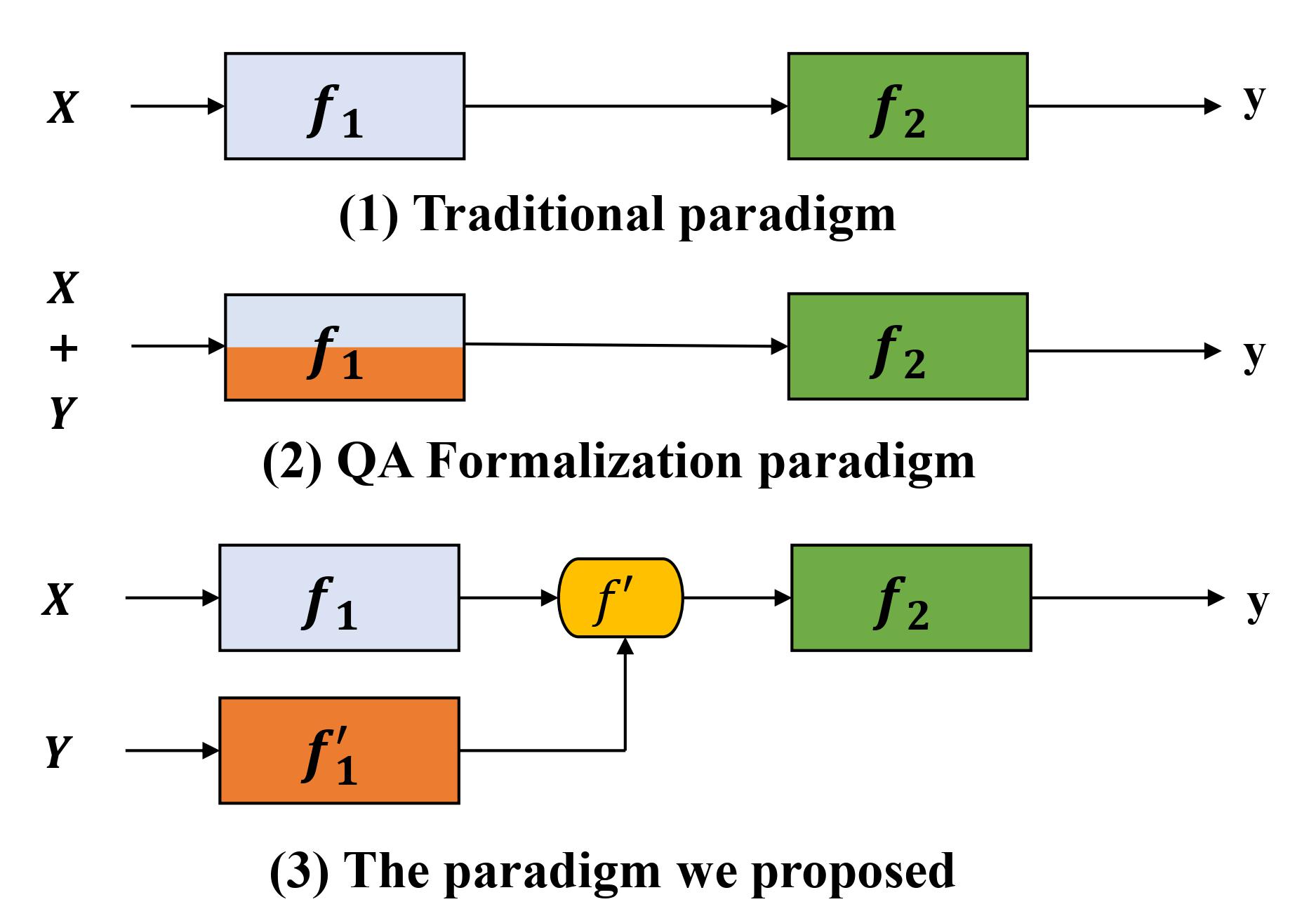}
	\caption{Illustration of different paradigms\protect\footnotemark \ for span extraction. $X$ represents the sequence; $Y$ represents the category-related extra input (e.g., question in QA paradigm); $y$ represents corresponding category; $f_1$ is the encoder to learning text representation; $f_2$ is the task layer to decode results; $f_1^{\prime}$ is the extra encoder to learn the representation of $Y$; $f^{\prime}$ is the extra module for the fusion of text semantics and label knowledge.}
	\label{fig:three_paradigm}
\end{figure}

\footnotetext{The traditional paradigm represents the methods that ignore the label knowledge.}

Many efforts have been devoted to span extraction. Early approaches are mainly based on handcrafted features such as domain dictionaries~\citep{sekine2004definition,etzioni2005unsupervised} and lexical features~\citep{ahn2006stages}. As neural networks show the effectiveness of learning text features automatically, many neural-based methods have been
proposed~\citep{huang2015bidirectional,strubell2017fast,liu2018jointly,cui-etal-2020-edge}.
Recently, self-attention-based pre-trained language models such as BERT~\citep{devlin-etal-2019-bert} are widely used to boost the span extraction task~\citep{devlin-etal-2019-bert,yang-etal-2019-exploring}. However, most existing methods treat labels as independent and meaningless one-hot vectors, neglecting prior information of labels 
(referred to as label knowledge).   
\begin{figure}[h]
    \centering
    \includegraphics[width=0.48\textwidth]{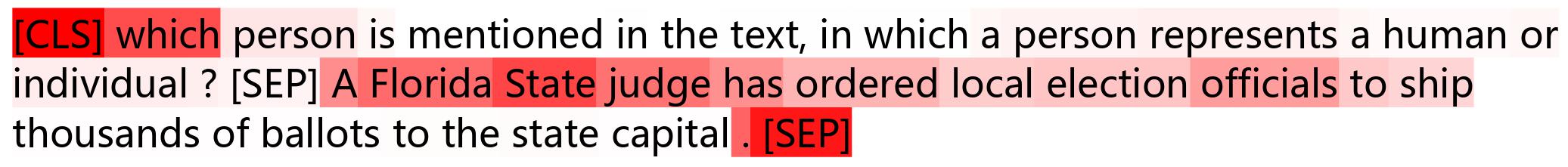}
    \caption{The Visualization of attention mechanism for the token "judge" (QA Formalization).\protect\footnotemark\   The darker color indicates the higher attention score.}
    \label{fig:qa_attn_vis}
\end{figure}

\footnotetext{The attention score comes from the well-trained model based on the pervious work~\citep{li-etal-2020-unified}.}


To alleviate the limitation, several studies  ~\citep{wang-etal-2018-label,lin-etal-2019-cost,chen-etal-2020-reading} start to integrate label knowledge into span extraction.
Among them, QA Formalization is especially attractive due to its effectiveness~\citep{levy-etal-2017-zero,li-etal-2019-entity,li-etal-2020-unified, liu-etal-2020-event, du-cardie-2020-event,li-etal-2020-event}. Simply put, QA Formalization treats span extraction as a question answering problem. 
Taking NER as an example, to extract ``PERSON'' entities, it is formalized as answering the question \textit{``which person is mentioned in the text, in which a person represents a human or individual?"} based on the given text. Benefiting from the label knowledge of the category-related questions, QA Formalization usually yields state-of-the-art performance in span extraction even in low-resource scenarios. 

QA Formalization, however, exhibits two key weaknesses: 
1) \textbf{Inefficiency}: Formalizing the span extraction as QA causes a drastic reduction of training/inference efficiency. Specifically, the typical QA Formalization method concatenates \textit{question} and \textit{text} as the input (e.g., [CLS] \textit{question} [SEP] \textit{text} [SEP]) and jointly encodes \textit{question} and \textit{text} with a transformer-based encoder. The joint-encoding has to transform every \textit{text} into $|C|$ pairs of the form $\langle \textit{question},\textit{text} \rangle$, where $|C|$ is the size of the label category set. This transformation, which increases both the size of the sample set and the length of text sequences, finally increases the time cost of training and inference. 
%
2) \textbf{Underutilization}: The label knowledge is integrated implicitly into text representation based on the self-attention mechanism~\citep{vaswani2017attention}. As Figure~\ref{fig:qa_attn_vis} shows, 
the ``attention'' of self-attention mechanism will be distracted by \textit{text}, not entirely focus on the \textit{question} part.
%
Thus, the label knowledge is not fully exploited to enhance the text representations.


To address aforementioned two problems, we propose a novel paradigm (seen in Figure~\ref{fig:three_paradigm}) to integrate label knowledge.
%
%
First, since joint-encoding causes low efficiency, we decompose \textit{question}-\textit{text} encoding process into two separate encoding modules: the \textit{text} encoding module $f_1$ and the \textit{question} encoding module $f^\prime_1$. In this way, the size of the sample set is no longer expanded by $|C|$ times.
%
%
Second, to fully utilize the label knowledge, a fusion module $f'$ is designed to explicitly integrate the label and the text representations.

To instantiates the above paradigm, we further propose a model termed as \textbf{LEAR} to learn \underline{L}abel-knowledge \underline{E}nh\underline{A}nced \underline{R}epresentation.
A powerful encoder $f^{\prime}_1$ is essential for understanding the \textit{label annotations}\footnote{Previous SOTA QA Formalization method~\citep{li-etal-2020-unified} adopts the annotations of label as \textit{questions}.}.
However, training the encoder $f_1^{\prime}$ from scratch is challenging since the number of \textit{label annotations} is too small. Thus we share the weights of $f_1$ and $f_1^{\prime}$~(called \textit{shared encoder}), which can learn the label knowledge by large pre-trained model and does not introduce extra parameters. Next, the learned label knowledge is integrated into text representations by the semantics-guided attention module.
%
We conduct experiments in five benchmarks on three typical span extraction tasks: flat NER, nested NER, and event detection~(ED).
Compared with QA Formalization baselines, our model LEAR outperforms them to achieve a new state-of-the-art. Furthermore, LEAR reduces training time and inference time by 76\% and 77\% on average, respectively.





To sum up, our contributions are as follows:
\begin{itemize}
	\item We propose a new paradigm to exploit label knowledge to boost span extraction, which encodes \textit{texts} and \textit{label annotations} independently and integrates label knowledge into text representation explicitly. 
	\item We propose a novel model, LEAR, to instantiate the above paradigm. It designs the \textit{shared encoder} and semantics-guided attention to tackle the technical challenges.
	\item The experiments show that our method achieves SOTA performance on four benchmarks, and it is much faster than the previous SOTA approach. Further analysis confirms the effectiveness and efficiency of our model.
\end{itemize}

\section{Preliminaries}
\subsection{Task Formalization}\label{sec:task_formalization}

\begin{figure*}[!h] 
\centering 
\includegraphics[width=0.90\textwidth]{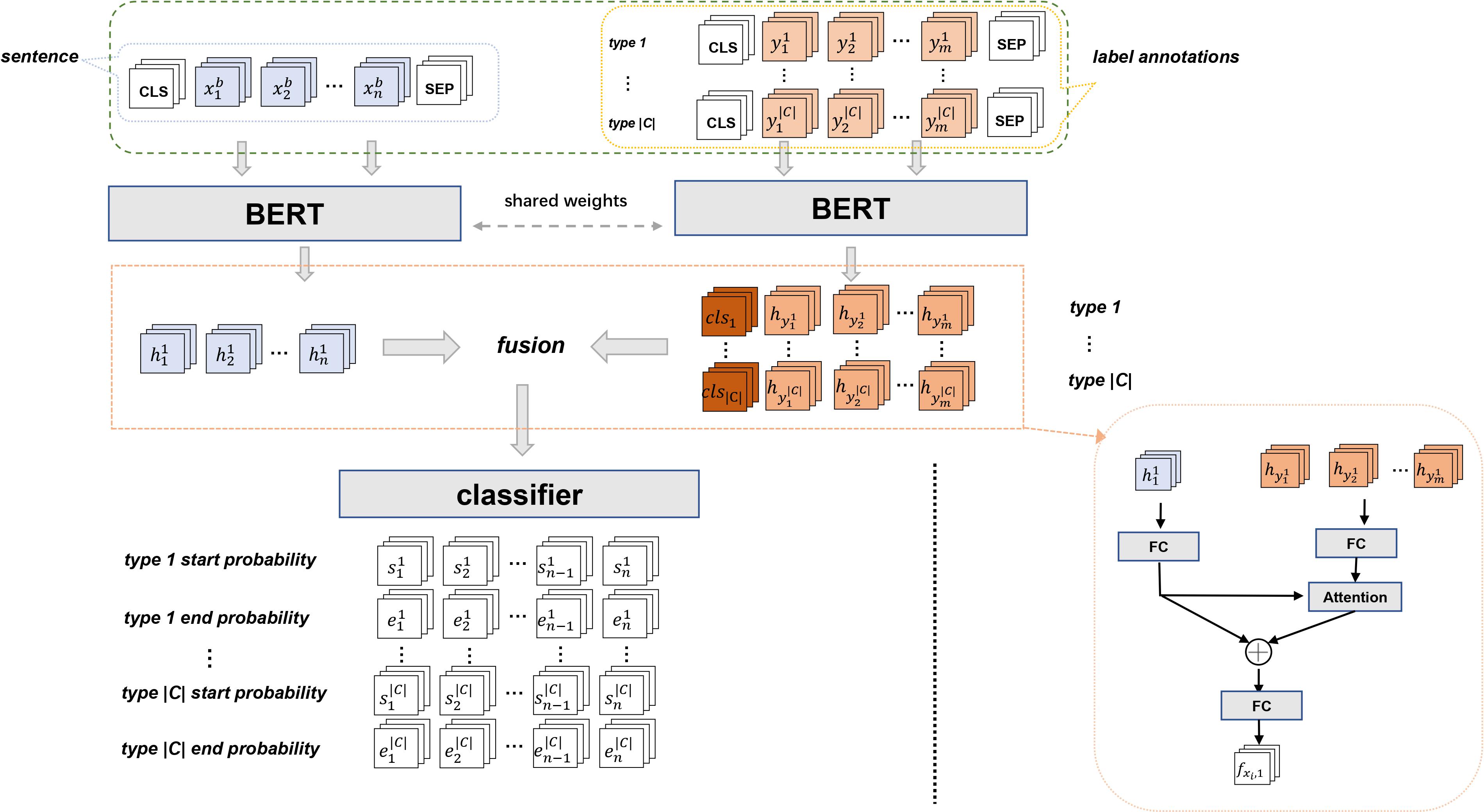} 
\caption{Illustration of our LEAR.} 
\label{Fig.1 model} 
\end{figure*}

We formulate the following span extraction task: given an input text $X=(x_1, x_2, \cdots, x_n)$ consisting of $n$ tokens, find out all candidate spans in $X$ and assign a label $c\in C$ to each of them, where $C$ is a predefined set of categories (or tag types, interchangeably). 
%


This formulation provides a uniform framework for modeling many important problems. For example, when $C$ is the set of event types such as die, attack, marry, and so on, span extraction is exactly the event detection task. In addition, if $C$ consists of entity types such as persons, organizations, locations, span extraction turns into the well-known named entity recognition task.



\subsection{Data Construction}
\begin{table}[!htp]\small
\centering
\begin{tabular}{ccc}
\toprule 
\textbf{Task} & \textbf{Category} & \textbf{Label Annotation}                                                             \\ \midrule
 ED & Die & \begin{tabular}[c]{@{}l@{}}a die Event occurs whenever the \\life of a person entity ends.\end{tabular}                                                   \\ \hline
NER & Person & \begin{tabular}[c]{@{}l@{}}a person entity is limited to human\\ including a single individual or a \\group. \end{tabular} \\       
 \bottomrule
\end{tabular}
\caption{Label categories and their corresponding annotations.}
\label{ace05-data-annotation}
\end{table}

QA formalization is powerful in span extraction since it incorporates label knowledge. One of its prerequisites is the existence of reasonable \textit{questions}. Usually, \textit{questions} are generated by a manually-designed pre-processing step, which is costly and lacks versatility and accessibility. For instance,~\citet{du-cardie-2020-event} and~\citet{li-etal-2020-unified} use a purpose-designed template to generate questions, while~\citet{liu-etal-2020-event} exploits a well-designed large pretraining model.

Previous work\footnote{\textit{Questions} are available in their open source project.}~\citep{li-etal-2020-unified} on flat and nested NER uses the annotations of each category~(referred to as \textit{label annotations}) as the \textit{questions}. We follow this setting in our work for a fair comparison. Similarly, we utilize the annotations of event types in ACE 2005 event detection task\footnote{All label annotations  are available at: \url{https://www.ldc.upenn.edu/sites/www.ldc.upenn.edu/files/english-events-guidelines-v5.4.3.pdf}.}. Table~\ref{ace05-data-annotation} presents an example of those annotations.



\section{Approach}
In this section, we first give an overall description of our LEAR architecture. LEAR consists of three crucial modules: semantics encoding module, semantics fusion module, and span decoding module. Our architecture (Figure \ref{Fig.1 model}) takes text $X$ and label annotation $Y$ of category set $C$ as input. 
%
The two inputs are respectively processed by two encoder networks whose backbone is BERT~\citep{devlin-etal-2019-bert}.
The two encoders share weights~(referred to as \textit{shared encoder}) while processing the two inputs. 
Then the text embedding and label embedding produced by the \textit{shared encoder} are fused by the semantic fusion module to derive the label-knowledge-enhanced embeddings for the text. Finally, the label-knowledge-enhanced embeddings are used to predict whether or not each token is a start or end index for some category.

\subsection{Semantics Encoding Module}

Semantics encoding module aims to encode the text and the label annotation into real-valued embeddings.
%
Since the number of label annotations is small compared with the whole sample set, it is challenging to build an encoder from scratch for the label annotations. Thus we introduce the shared encoder, which is inspired by siamese networks~\citep{bromley1993signature}. The shared encoder is efficient in learning the representation of label annotations and does not introduce extra parameters.  
%


Given input text $X$ and label annotations $Y$, LEAR first extracts their embeddings $h_X\in {\mathcal{R}^{n\times d}}$ and $h_Y\in {\mathcal{R}^{|C|\times m\times d}}$, where $n$ is the length of $X$, $m$ is the length of label annotation, $|C|$ is the size of the category set $C$, and $d$ is the vector dimension of the encoder. 
%
We denote this operation as:
\begin{align}
& h_X = f_1(X) \\
& h_Y = f_1^{\prime}(Y)
\end{align}


\subsection{Semantic Fusion}

The semantic fusion module aims at enhancing the text representation with label knowledge explicitly. 
To this end, we devise a semantics-guided attention mechanism to achieve this goal.

Specifically, we first feed $h_{X}$ and $h_Y$ into a fully connected layer, respectively, to map their representations into the same feature space:
\begin{align} 
	h^{\prime}_{X} & = \mathbf{U}_{1}\cdot h_{X} \\
	h^{\prime}_{Y} & = \mathbf{U}_{2}\cdot h_{Y} 
\end{align}
where $\mathbf{U}_1,\mathbf{U}_2 \in \mathcal{R}^{d\times d}$ be the learnable parameters of the fully connected layers.

Then, we apply the attention mechanism over the label annotations for each token in the text.
For any $ 1\le i\le n$, let $x_i$ be the $i$th token of $X$, and $h^{\prime}_{x_i}\in\mathcal{R}^d$ be the $i$th row of $h^{\prime}_{X}$. 
Likewise, for any $1\le i\le m$ and category $c\in C$, let $y_{j}^{c}$ be the $j$th token of the annotation of $c$, and $h^{\prime}_{y_{j}^{c}}$ be its embedding from $h^{\prime}_{Y}$. 
%
We compute the dot product of $h^{\prime}_{x_i}$ and $h^{\prime}_{y_{j}^{c}}$, and apply a softmax function to obtain the attention scores:
\begin{equation}
    a_{x_{i}, y_{j}^{c}} = \frac{\exp \left(h_{x_{i}}^{\prime} \cdot h^{\prime}_{y_{j}^{c}} \right)}{\sum_{j} \exp \left(h_{x_{i}}^{\prime} \cdot h^{\prime}_{y_{j}^{c}}\right)}
\end{equation}
 
Finally, we get the fine-grained features by attention,
which is in turn fused into token embedding by \textit{add} operation:
\begin{align}
 h_{x_{i}}^{c} &= h^{\prime}_{x_{i}} +  \sum_{j}  a_{x_{i}, y_{j}^{c}}\cdot h^{\prime}_{y_{j}^{c}} \\
 \hat{h}_{x_{i}}^{c} &= \tanh{(\mathbf{V}\cdot h^c_{x_{i}} + \mathbf{b})}
\end{align}
where $\tanh{(\cdot)}$ is the hyperbolic tangent function, and $\mathbf{V} \in \mathcal{R}^{d\times d}$ and $\mathbf{b} \in \mathcal{R}^{d}$ are learnable parameters. 
Intuitively, $\hat{h}^c_{x_{i}}$ encodes the information related to category $c$.


Repeating the process for all categories, we obtain  
the category-related embedding $\hat{h}_{x_i}=(\hat{h}^1_{x_{i}}, \cdots, \hat{h}^{|C|}_{x_{i}})$ for each token $x_{i}$. 

\subsection{Span Decoding}
\label{section. span decoding}
Now we are ready to select spans. 
Following~\citet{li-etal-2020-unified}, we use the start/end tagging schema to annotate the target spans to extract.
Specifically, for each token $x_i$, we compute the following vector:
\begin{align}\label{eq:start}
\text{start}_{x_{i}}= \operatorname{sigmoid}(f_{o}(\mathbf{M_{s}} \circ \hat{h}_{x_{i}} + \mathbf{b_s}))
\end{align}
where $\mathbf{M_{s}} \in \mathcal{R}^{|C| \times d}$ and $\mathbf{b_s}  \in \mathcal{R}^{d}$ are learnable parameters, $\circ$ is the element-wise multiplication, and $f_{o}(\cdot)$ is the function that sums up the rows of the input matrix. Intuitively, for any $c\in C$, $\text{start}^c_{x_{i}}$ indicates the probability that $x_i$ starts a span of the category $c$.

Likewise, we obtain the $\text{end}_{x_{i}}$, which indicates the probabilities that $x_i$ ends a span, in the same prediction procedure.
%
%
Then we extract the results case by case, depending on whether or not spans of the same category can be nested\footnote{\textit{Nested} here represents both nested and overlapped spans, just like nested NER \citep{finkel-manning-2009-nested}.}. 

\paragraph{Flat Span Decoding} This is the case without nested spans in the same category.

The most widely adopted method is the \textit{nearest matching principle} \citep{du-cardie-2020-event,wei-etal-2020-novel}, which matches a start position of category $c$ with the nearest next end position of $c$. 

In contrast, we follow the \textit{heuristic matching principle}~\citep{yang-etal-2019-exploring-pre}, which determines spans from the lens of probability. Roughly speaking, among candidate start and end positions of a category $c$, we only match those having high probabilities, where the probabilities are derived from vectors defined in formulas (\ref{eq:start}) 
For detailed information of heuristic matching, please refer to the algorithm in Appendix~\ref{appendix:Heuristic Match Principle}.


The two principles for span decoding are further compared by experiments in Appendix~\ref{section:span_decoding_strategy}.

\paragraph{Nested Span Decoding} Now suppose that spans in the same category may be nested or overlapped.

Since the \textit{heuristic matching principle} does not work anymore, we follow the solution of BERT-MRC \citep {li-etal-2020-unified}. It employs a binary classifier to predict the probability that a pair of candidate start/end positions should be matched as a span. Specifically, for any category $c$, 
define the following binary classifier:
\begin{align}\label{eq:finalprob}
 P^{c}_{i,j} = \operatorname{sigmoid}(\mathbf{M}\cdot\operatorname{concat}(\hat{h}^c_{x_i},\hat{h}^c_{x_j}) ) 
\end{align}
where $1\le i,j\le n$, and $\mathbf{M}\in \mathcal{R}^{1\times2d}$ is the learnable parameter.
When $P^{c}_{i,j}>0.5$, it will be predicted that $x_{i}$ and $x_{j}$ demarcate a span of $c$.

\subsection{Loss Function}
\label{section. loss}

Given input text $X=(x_1, x_2, \cdots, x_n)$ consisting of $n$ tokens and set $C$ of categories, for any $c\in C$, define $S^{c}\in \{0,1\}^n$ to be the vector whose $i$th entry $S_{x_i}^{c}=1$ if and only if $x_i$ is a ground-truth start position of $c$. Likewise, define $E^{c}\in \{0,1\}^n$ to indicate the ground-truth end positions. Recall the vectors $\text{start}^{c}$ and $\text{end}^{c}$ defined in Section~\ref{section. span decoding}. Define start loss function $\mathcal{L}_{s}$ and end loss function $\mathcal{L}_{e}$ of our model as follows:
$$
\mathcal{L}_{s}=\frac{1}{n} \sum_{c \in C}\sum_{1\le i\le n} \operatorname{CE}\left(\text{start}_{x_i}^{c},S_{x_i}^{c}\right)
$$
$$
\mathcal{L}_{e}=\frac{1}{n} \sum_{c \in C}\sum_{1\le i\le n} \operatorname{CE}\left(\text{end}_{x_i}^{c},\ E_{x_i}^{c}\right)
$$
where CE stands for the cross entropy.

\paragraph{Flat Span Extraction} The final loss function of our model is defined to be $\mathcal{L}=\mathcal{L}_{s}+\mathcal{L}_{e}$.

\paragraph{Nested Span Extraction} 
More notation is needed. Recall the matrix $P^c\in \mathcal{R}^{n\times n}$ defined in Formula (\ref{eq:finalprob}). Let $M^{c}\in \mathcal{R}^{n\times n}$ be the binary matrix such that $M^{c}_{i,j}=1$ if and only if the tokens $x_{i}$ and $x_{j}$ demarcate a ground-truth span of category $c$. Define the match loss function
$$
\mathcal{L}_{match}=\frac{1}{n^2}\sum_{1\le i,j\le n}\sum_{c \in C} \operatorname{CE}\left(P^{c}_{i,j},M^{c}_{i,j}\right) W^{c}_{i,j}
$$
where $W^{c}\in \mathcal{R}^{n\times n}$ is the binary matrix such that $W^{c}_{i,i}=1$ if and only if  $P^c_{i,j}>0.5$ or $M^{c}_{i,j}=1$.

Then the final loss function of our model is defined to be $\mathcal{L}=\alpha(\mathcal{L}_{s}+\mathcal{L}_{e})+\beta\mathcal{L}_{match}$, where $\alpha,\beta$ are hyper-parameters to control the contributions towards the overall training objective.

\section{Experiments}
In this section, we present LEAR results on 5 widely-used benchmarks.

\subsection{Datasets}

\paragraph{Dataset} We evaluate our model on three span extraction tasks: flat NER, nested NER and event detection.
For flat NER, we conduct experiments on MSRA~\citep{levow-2006-third} and Chinese OntoNote 4.0~\citep{pradhan-etal-2011-conll}.
For nested NER, we evaluate our model on ACE 2004~\citep{doddington-etal-2004-automatic} and ACE 2005~(NER) datasets.
For event detection, we use the ACE 2005\footnote{This corpora is designed for multi-tasks, such as event detection and NER. Data source: \url{https://catalog.ldc.upenn.edu/LDC2006T06}}~(ED) dataset.

For MSRA and Chinese OntoNote 4.0, which contains three and four types of entities respectively, we follow the data preprocessing strategies in~\citet{li-etal-2020-unified} and~\citet{meng2019glyce} for fair comparison. 
ACE 2005~(NER) and ACE 2004 both annotate 7 entity categories. For ACE 2005~(NER), we use the same data split as previous works~\citep{lin-etal-2019-sequence}; for ACE 2004, We use the same setup as~\citet{katiyar-cardie-2018-nested}. 
ACE 2005~(ED) annotates 33 types of events and we follow the same settings of~\citet{chen-etal-2015-event} and~\citet{chen-etal-2018-collective} to split data into train, development, and test set.
More statistics of datasets are listed in Appendix~\ref{sec:data_statistics}.






\subsection{Baselines}





\paragraph{Named Entity Recognition} We use the following models as baselines: 
(1) \textbf{BiLSTM-CRF} \citep{ma-hovy-2016-end} uses the Bi-LSTM layer as encoder. (2) \textbf{Seg-Graph} \citep{wang-lu-2018-neural} proposes a segmental hypergragh representation to model overlapping entity mentions. 
(3) \textbf{BERT-Tagger} \citep{devlin-etal-2019-bert} treats NER as a tagging task with a bidirectional encoder representations.
(4) \textbf{Lattice-LSTM} \citep{zhang-yang-2018-chinese} constructs a word-character lattice for Chinese NER. 
(5) \textbf{Glyce-BERT} \citep{meng2019glyce} combines glyph information with BERT pretraining for Chinese NER. 
(6) \textbf{Seq2Seq-BERT} \citep{shibuya-hovy-2020-nested} views the nested NER as a sequence-to-sequence problem. 
%
%
(7) \textbf{Biaffine-NER}~\citep{yu-etal-2020-named} predicts named entity with a biaffine network.
(8) \textbf{BERT-MRC} \citep{li-etal-2020-unified} treats NER as a MRC/QA task, which is the state-of-the-art method on both flat and nested NER. 

\paragraph{Event Detection} We compare with the following methods: (1) \textbf{DMCNN} \citep{chen-etal-2015-event} builds a dynamic multi-pooling convolutional model;
(2) \textbf{JRNN} \citep{nguyen-etal-2016-joint} employs bidirectional RNN for ED; 
(3) \textbf{ANN-AugAtt} \citep{liu-etal-2017-exploiting} uses annotated event argument information to get better attention scores; 
(4) \textbf{JMEE} \citep{liu-etal-2018-jointly} enhances GCN with self-attention and highway network;
(5) \textbf{EE-GCN} \citep{cui-etal-2020-edge} learns token representation via edge-enhanced GCN with specific syntactic label incorporated.
(6) \textbf{EKD} \citep{tong-etal-2020-improving} is the state-of-the-art method on the ACE2005 dataset. 
(7) \textbf{BERT\_QA\_Trigger} \citep{du-cardie-2020-event} formalizes event detection as a QA task.

Furthermore, to compare the efficiency between QA Formalization and LEAR, we instantiate the traditional paradigm as a baseline for efficiency comparison in the simplest way, which only contains a BERT encoder and two fully connected layers as the classifiers.
We denote this baseline model as \textbf{Traditional Formalization}.

\subsection{Experimental Setups}

We use BERT~\citep{devlin-etal-2019-bert} as the backbone to learning the contextualized representation of the texts. More specifically, we implement our model based on the BERT-large model for NER task, which is the same as BERT-MRC~\citep{li-etal-2020-unified}. 
In the event detection task, we use the BERT-base model as the backbone. We adopt the adam optimizer~\citep{DBLP:journals/corr/KingmaB14} with a linear decaying schedule to train our model. The detail of hyper-parameters settings is listed in Appendix~\ref{appendix:hyper-parameters}.

To make results comparable in the efficiency comparison experiment~(as shown in Table~\ref{tab:efficiency comparison}), all models take the BERT-base as the backbone and set all hyperparameters to the same except \textit{max\_seq\_len} of QA Formalization. The higher \textit{max\_seq\_len} meets the requirement of taking the \textit{question} as extra input for QA Formalization. 


\paragraph{Effectiveness Evaluation} We use micro-average precision, recall, and F1 as evaluation metrics. A prediction is considered correct only if both its boundary and category are predicted correctly.

\paragraph{Efficiency Evaluation} We use the time costs (in seconds) of training and inference to evaluate the efficiency of different models. Specifically, 1) \textit{Training}: the time cost of training in one epoch; 2) \textit{Inference}: the time cost for the model to get all prediction results of the test set.

\begin{table}[!t]\scriptsize
\centering
\begin{tabular}{llll}
\toprule
\multicolumn{4}{c}{\textbf{English ACE2005 for ED (Flat)}}                        \\ \toprule
\textbf{Model} & \multicolumn{1}{l}{\textbf{P}} & \textbf{R} & \textbf{F1} \\
\hline 
DMCNN \citep{chen-etal-2015-event} & 75.6 & 63.6 & 69.1 \\
JRNN \citep{nguyen-etal-2016-joint}  & 66.0 & 73.0 & 69.3 \\
ANN-AugAtt \citep{liu-etal-2017-exploiting} & 78.0 & 66.3 & 71.7 \\
JMEE \citep{liu-etal-2018-jointly} & 76.3 & 71.3 & 73.7 \\
EE-GCN \citep{cui-etal-2020-edge} & 76.7 & 78.6 & 77.6 \\
BERT\_QA\_Trigger$^{\dagger}$ \citep{du-cardie-2020-event} & 71.12 & 73.70 & 72.39 \\
EKD \citep{tong-etal-2020-improving} & 79.1 & 78.0 & 78.6 \\
\midrule
LEAR & \textbf{82.04} & \textbf{81.18} & \textbf{81.61} \\
\toprule
\multicolumn{4}{c}{\textbf{English ACE 2004 for NER (Nested)}}                        \\ \toprule
\textbf{Model} & \multicolumn{1}{l}{\textbf{P}} & \textbf{R} & \textbf{F1} \\
\hline 
Seg-Graph \citep{wang-lu-2018-neural}   & 78.0 & 72.4 & 75.1 \\
Seq2seq-BERT \citep{strakova-etal-2019-neural} & - & - & 84.40 \\
DYGIE \citep{luan-etal-2019-general}   & - & - & 84.7 \\
BERT-MRC$^{\dagger}$ \citep{li-etal-2020-unified}  & 85.05 & \textbf{86.32} & 85.98 \\
Biaffine-NER~\citep{yu-etal-2020-named} & 87.3 & 86.0 & 86.7 \\
\midrule
LEAR     & \textbf{87.89} & 85.86 & \textbf{86.87}  \\
\toprule
\multicolumn{4}{c}{\textbf{English ACE 2005 for NER (Nested)}}                        \\ \toprule
\textbf{Model} & \multicolumn{1}{l}{\textbf{P}} & \textbf{R} & \textbf{F1} \\
\hline 
Seg-Graph \citep{wang-lu-2018-neural}   & 76.8 & 72.3 & 74.5 \\
DYGIE \citep{luan-etal-2019-general}   & - & - & 82.9 \\
Seq2seq-BERT \citep{strakova-etal-2019-neural} & - & - & 84.33 \\
Biaffine-NER~\citep{yu-etal-2020-named} & 85.2 & 85.6 & 85.4 \\
BERT-MRC$^{\dagger}$ \citep{li-etal-2020-unified}  & \textbf{87.16} & 86.59 & \textbf{86.88} \\
\midrule
LEAR     & 84.85 & \textbf{87.95} & 86.63  \\
\toprule
\multicolumn{4}{c}{\textbf{Chinese OntoNotes 4.0 for NER (Flat)}}                        \\ \toprule
\textbf{Model} & \multicolumn{1}{l}{\textbf{P}} & \textbf{R} & \textbf{F1} \\
\hline 
Lattice-LSTM \citep{zhang-yang-2018-chinese}  & 76.35 & 71.56 & 73.88 \\
BERT-Tagger \citep{devlin-etal-2019-bert}   & 78.01 & 80.35 & 79.16 \\
Glyce-BERT \citep{meng2019glyce} & 81.87 & 81.40 & 81.63 \\
BERT-MRC$^{\dagger}$ \citep{li-etal-2020-unified} \  & \textbf{82.98} & 81.25 & 82.11 \\
\midrule
LEAR     & 81.12 & \textbf{84.86} & \textbf{82.95}  \\
\toprule
\multicolumn{4}{c}{\textbf{Chinese MSRA for NER (Flat)}}                        \\ 
\toprule
\textbf{Model} & \multicolumn{1}{l}{\textbf{P}} & \textbf{R} & \textbf{F1} \\
\hline 
Lattice-LSTM \citep{zhang-yang-2018-chinese}  & 93.57 & 92.79 & 93.18 \\
BERT-Tagger \citep{devlin-etal-2019-bert}   & 94.97 & 94.62 & 94.80 \\
Glyce-BERT \citep{meng2019glyce} & 95.57 & 95.51 & 95.54 \\
BERT-MRC$^{\dagger}$ \citep{li-etal-2020-unified} \  & 96.18 & 95.12 & 95.75 \\
\midrule
LEAR & \textbf{96.23} & \textbf{95.57} & \textbf{95.96} \\
\toprule
\end{tabular}
\caption{Results in five benchmarks. The best results are in \textbf{bold}, $^{\dagger}$ means QA Formalization methods.}
\label{tab:span extraction results}
\end{table}
\begin{table*}[!htp]\scriptsize
    \centering
\begin{tabular}{@{}clcrrrrrr@{}}
        \toprule
        \multirow{2}{*}{\textbf{Task}} & \multirow{2}{*}{\textbf{Dataset}} & \multirow{2}{*}{$\mathbf{|C|}$} & \multicolumn{2}{c}{\textbf{Traditional Formalization}} & \multicolumn{2}{c}{\textbf{QA Formalization}} & \multicolumn{2}{c}{\textbf{LEAR}}                                                            \\ \cmidrule(l){4-9}
                                            &                                   &                                 & \textbf{Train}                                   & \textbf{Inference}                   & \textbf{Train}                    & \textbf{Inference} & \textbf{Train} & \textbf{Inference} \\ \cmidrule(l){1-9}
        ED                                  & ACE 2005                          & 33                              & 349.9(x1.0)                                            & 5.8(x1.0)                                  & 11456.2(x32.7)                    & 176.1(x30.4)       & 1005.5(x2.9)   & 9.8(x1.7)         \\\cmidrule(l){1-9}
        \multirow{4}{*}{NER}
                                            & MSRA                              & 3                               & 167.8(x1.0)                                            & 5.1(x1.0)                                  & 626.7(x3.7)                       & 14.5(x2.8)         & 206.8(x1.2)    & 5.6(x1.1)          \\
                                            & OntoNotes 4.0                     & 4                               & 58.3(x1.0)                                             & 5.2(x1.0)                                  & 258.1(x4.4)                       & 19.7(x3.8)         & 74.4(x1.3)     & 6.1(x1.2)          \\
                                            & ACE 2005                          & 7                               & 103.9(x1.0)                                            & 4.3(x1.0)                                  & 684.4(x6.6)

                                            & 26.3(x6.1)
                                            & 167.8(x1.6)                       & 5.1(x1.2)                                                                                                                                                                                                                \\
                                            & ACE 2004                          & 7                               & 87.1(x1.0)                                             & 3.4(x1.0)                                  & 604.8(x6.9)                       & 20.5(x6.0)         & 145.5(x1.7)    & 4.3(x1.3)          \\ \hline
    \end{tabular}
    \caption{The efficiency comparison of different methods. 
    %
    $( \cdot )$ indicates the relative efficiency compare with the Traditional Formalization (e.g., $\frac{T_{LEAR}}{T_{Traditional\ Formalization}}$ ).  
    %
    %
	}
    \label{tab:efficiency comparison}
\end{table*}


\subsection{Main Results}

\paragraph{Effectiveness} Table~\ref{tab:span extraction results} shows the performance of our LEAR compared with the above state-of-the-art methods on the test sets. We can see that our LEAR outperforms all other models on four benchmarks, i.e., +3.01\%, +0.84\%, +0.21\%, +0.17\%, respectively on ACE 2005 (ED), OntoNote 4.0, MSRA and ACE 2004. 
This improvement indicates that the \textit{explicit} fusion with a dedicated module is better than the \textit{implicit} fusion based on the self-attention mechanism. 
Since the joint-encoding of QA Formalization, the ``attention'' of self-attention mechanism will be distracted by \textit{text}, not entirely focus on the \textit{question}. Thus the label knowledge introduced by \textit{label annotation} is not fully exploited. By contrast, our LEAR learns knowledge-enhanced representations for each token by a semantics-guided fusion module, whose attention entirely focuses on the \textit{label annotation}.



\paragraph{Efficiency} 

Table~\ref{tab:efficiency comparison} shows that our LEAR is much faster than QA Formalization, i.e., reducing the training and inference time by 76\% and 77\% on average, respectively. The reduction in training/inference time is positively correlated with the number of categories $|C|$, which benefits from breaking the joint-encoding limitation of QA Formalization. As Table~\ref{tab:complexity} shows, the time complexity of LEAR during inference is $\mathcal{O}(n^2+|C|mn)$, in which we ignore the cost for the encoding of label annotations in our LEAR. Because LEAR only encodes all label annotations once and reuses their representations during the inference, which is favorable for industrial applications in the resource-limited online environment. In contrast, the time complexity of QA Formalization is $\mathcal{O}(|C| \cdot (n+m)^2)$, causing a dramatic decrease in efficiency of inference.






\begin{table}[!h]\small
\centering
\begin{tabular}{lc}
\hline
\textbf{Method} & \textbf{Time Complexity} \\ \hline
Traditional Formalization         &    $\mathcal{O}(n^2)$  \\
QA Formalization        &   $\mathcal{O}(|C| \cdot (n+m)^2)$               \\
LEAR       &  $\mathcal{O}(n^2+|C|mn)$  \\ \hline
\end{tabular}
\caption{The time complexity of different model architectures during inference.}
\label{tab:complexity}
\end{table}



To summarize, the fundamental starting points of the proposed paradigm include: 1) decomposing question-text joint encoding into two separate encoding modules; 2) explicitly integrating label knowledge by a dedicated module. The above experiments confirm that our LEAR, an instantiation of the proposed paradigm, outperforms previous SOTA methods in \textbf{effectiveness and efficiency}.

\section{Analysis}
\subsection{Analysis for Model Variants}

\begin{table*}[htp]\small
\centering
\begin{tabular}{@{}llllll@{}}
\toprule
\textbf{Model} & \multicolumn{1}{l}{\textbf{ACE 2005 (ED)}} & \multicolumn{1}{l}{\textbf{ACE 2005 (NER)}} & \multicolumn{1}{l}{\textbf{ACE 2004}} & \multicolumn{1}{l}{\textbf{OntoNotes 4.0}} & \multicolumn{1}{l}{\textbf{MSRA}} \\ \midrule
\textbf{LEAR} & \textbf{81.61} & \textbf{86.63} & \textbf{86.87} & \textbf{82.95} & \textbf{95.96} \\
\ \ -- LNE & 80.96~($\downarrow0.65$) & 85.83~($\downarrow0.80$) & 86.20~($\downarrow0.67$) & 82.58~($\downarrow0.37$) & 95.82~($\downarrow0.14$) \\
\ \ -- LEL & 79.72~($\downarrow1.89$)  & 85.34~($\downarrow1.29$) & 85.88~($\downarrow0.99$) & 82.92~($\downarrow0.03$) & 95.85~($\downarrow0.11$) \\
\ \ -- AP \& Add & 79.68~($\downarrow1.93$)  & 85.74~($\downarrow0.89$) & 86.18~($\downarrow0.69$) & 82.56~($\downarrow0.39$) & 95.88~($\downarrow0.08$)  \\
\ \ -- SF \& Sim & 79.76~($\downarrow1.85$)  & 86.31~($\downarrow0.32$)  & 86.67~($\downarrow0.2$) & 82.72~($\downarrow0.23$) & 95.76~($\downarrow0.2$) \\ \bottomrule
\end{tabular}
\caption{The performance of the model variants. The values in table are F1 scores on test sets.}
\label{tab:ablation_analysis}
\end{table*}

%
To demonstrate the effectiveness of our method, we build a series of variants of LEAR. For the semantics encoding module, we set: 
1) \textbf{Label Embedding Layer~(LEL)}: replacing the encoder module of label annotations with a label embedding layer, which is initialized by glove~\citep{pennington2014glove}. The F1 scores drop 0.86\% on average. The results show that the improvement of our LERA comes from understanding the label annotation, which is handled well by the shared encoder.
2) \textbf{Label Name Encoding~(LNE)}: replacing the label annotations with corresponding label names. The results drop 0.53\% on average, indicating that label names contain less label knowledge than label annotation.

In order to survey the semantics fusion strategy, we set: 
1) \textbf{Average Pooling \& Add ~(AP \& Add)}: replacing the semantics-guided attention mechanism with average pooling and integrating label knowledge by \textit{add} operation. The F1 scores drop by 0.80\% on average. 
2) \textbf{Sentence Features \& Similarity~(SF \& Sim)}: using the sentence-level features of label annotations (i.e., the embedding of [CLS] symbol) instead of token-level features. Thus the semantics-guided attention mechanism turns into the similarity calculation between token embedding and label feature.
The F1 scores drop by 0.56\%.
The above two settings retain the extra learnable parameters introduced by the fusion module. The results show that the improvement comes from the better exploitation of label knowledge, not the larger parameters. Besides, the results demonstrate that fine-grained~(i.e., token-level) features are more effective. 

All the above experiments show the effectiveness of our LEAR. Furthermore, the worst-performing variants of LERA still rival the QA Formalization method, which powerfully demonstrates the superiority of the proposed paradigm.




\subsection{Performance in Data-Scarce Scenarios}
\begin{table}[!h]\scriptsize
\centering
\begin{tabular}{@{}cccc@{}}
\toprule
\textbf{Dataset}                 & \textbf{Settings} & \textbf{LEAR$_{\text{w/o}}$} & \textbf{LEAR} \\ \midrule

\multirow{2}{*}{ACE 2005 (NER)} & 1-shot            &   3.23  &   \textbf{15.42}            \\
                                & 5-shot            &   38.77   &   \textbf{43.92} \\\cmidrule(){1-4}
\multirow{2}{*}{ACE 2004}       & 1-shot            &  13.30                &   \textbf{22.81}            \\
                                & 5-shot            &   38.11               &    \textbf{39.03}          \\\cmidrule(){1-4}
\multirow{2}{*}{OntoNotes 4.0}  & 1-shot            &   1.89            &   \textbf{7.28}   \\
                                & 5-shot            &   39.21           &   \textbf{41.32}             \\\cmidrule(){1-4}
\multirow{2}{*}{MSRA}           & 1-shot            &   0.16             & \textbf{0.39}       \\
                                & 5-shot            &   21.28            &  \textbf{26.22}   \\\cmidrule(){1-4} 
\multirow{2}{*}{ACE 2005 (ED)}  & 1-shot            &  23.31                & \textbf{30.23} \\ 
                                & 5-shot            &  63.04                &  \textbf{63.52}   \\ 
                                \bottomrule
\end{tabular}
\caption{F1 scores on exploring the extremely data-scarce scenarios.\protect\footnotemark \ Both methods take the BERT-base as the base model. The best results are in \textbf{bold}.}
\label{tab:fewshot}
\end{table}

To verify that exploiting label knowledge is beneficial in data-scarce scenarios, we introduce LEAR$_{\text{w/o}}$ for comparison. LEAR$_{\text{w/o}}$ is short for LEAR without label knowledge, whose settings are the same with LEAR except that BERT alone rather than shared encoder and label semantic fusion module are used~(i.e., the standard fine-tuning). We conduct two sets of experiments for each dataset using various proportions of the training data: 1-shot and 5-shot. For the 1-shot setting, we sample one sentence for each category in the training set, and the setting of 5-shot is similar. We repeat each experiment 5 times. Tabel \ref{tab:fewshot} shows that our LEAR demonstrates superior performance, for example, obtaining up to +12\% absolute improvement and +6.8\% on average across all datasets in the 1-shot setting. This is in line with our expectation since LEAR enhances the text representation with label knowledge, which provides more prior information. 

\footnotetext{We does not compare with QA paradigm methods because prior works does not report their training data.}

In the appendix, we list the further analysis about the effect of different span decoding strategies and the comparison between solving span extraction in the multi-label classification (our LEAR) or sequence-labeling manner (e.g., a CRF layer).

\section{Related Work}

\paragraph{Event Detection (ED).} Event Detection aims at extracting event triggers from a text and classifying them. It is dominantly solved in a representation-based manner, where triggers are represented by embedding. In case of no extra information, the representation can be obtained by a powerful text encoder which is usually based on CNN~\citep{chen-etal-2015-event}, RNN~\citep{nguyen-etal-2016-joint}, or attention mechanism~\citep{yang-etal-2019-exploring-pre,tong-etal-2020-improving}.
Besides, the representation can be enhanced by extra information.
Examples of typical extra information include syntactic information~\citep{liu-etal-2018-jointly,cui-etal-2020-edge} and knowledge base~\citep{liu-etal-2016-leveraging,chen-etal-2017-automatically}. In particular, label knowledge is attracting more and more attention~\citep{li-etal-2020-event,du-cardie-2020-event}, which usually formalizes ED as a QA problem.




\paragraph{Named Entity Recognition (NER).} 
Named entity recognition seeks to locate named entities in an unstructured text and classify them into pre-defined categories such as person, organization, location, etc. Traditional methods treat it as a classification task and use CRFs~\citep{lafferty2001conditional,sutton2007dynamic} as the backbone. 
Then neural networks become a prevalent tool in NER with the development of deep learning. Recently, the performance of NER has been further improved by large-scale language models such as ELMo~\citep{peters-etal-2018-deep} and BERT~\citep{devlin-etal-2019-bert}. When label knowledge is available, state-of-the-art performance can be obtained by formulating NER as a QA problem.


\section{Conclusion}



In this paper, we propose a novel paradigm to exploit label knowledge to boost the span extraction task and further instantiate a model named LEAR.
Unlike the existing QA Formalization methods, LEAR first encodes the text and label annotations independently, and uses a semantic fusion module to integrate label knowledge into the text representation explicitly. 
In this way, we can overcome the \textit{inefficiency} and \textit{underutilization} problems of QA Formalization.
Experimental results show that our model outperforms the previous works and enjoys a significantly faster training/inference speed.

\section*{Acknowledgments}
We would like to thank all reviewers for their insightful comments and suggestions. This work is partially supported by Key-Area Research and Development Program of Guangdong Province (NO.2020B010164003), the National Natural Science Foundation of China (62072433,62090020), the Fundamental Research Funds for the Central Universities (No. DUT21LAB302), Youth Innovation Promotion Association of Chinese Academy of Sciences (2013073), and the Strategic Priority Research Program of Chinese Academy of Sciences (Grant No. XDC05030200).
\bibliographystyle{acl_natbib}
\bibliography{mybib}

\begin{thebibliography}{49}
\expandafter\ifx\csname natexlab\endcsname\relax\def\natexlab#1{#1}\fi

\bibitem[{Ahn(2006)}]{ahn2006stages}
David Ahn. 2006.
\newblock The stages of event extraction.
\newblock In \emph{Proceedings of the Workshop on Annotating and Reasoning
  about Time and Events}, pages 1--8.

\bibitem[{Bromley et~al.(1993)Bromley, Bentz, Bottou, Guyon, LeCun, Moore,
  S{\"a}ckinger, and Shah}]{bromley1993signature}
Jane Bromley, James~W Bentz, L{\'e}on Bottou, Isabelle Guyon, Yann LeCun, Cliff
  Moore, Eduard S{\"a}ckinger, and Roopak Shah. 1993.
\newblock Signature verification using a “siamese” time delay neural
  network.
\newblock \emph{International Journal of Pattern Recognition and Artificial
  Intelligence}, 7(04):669--688.

\bibitem[{Chen et~al.(2017)Chen, Liu, Zhang, Liu, and
  Zhao}]{chen-etal-2017-automatically}
Yubo Chen, Shulin Liu, Xiang Zhang, Kang Liu, and Jun Zhao. 2017.
\newblock \href {https://doi.org/10.18653/v1/P17-1038} {Automatically labeled
  data generation for large scale event extraction}.
\newblock In \emph{Proceedings of the 55th Annual Meeting of the Association
  for Computational Linguistics (Volume 1: Long Papers)}, pages 409--419,
  Vancouver, Canada. Association for Computational Linguistics.

\bibitem[{Chen et~al.(2015)Chen, Xu, Liu, Zeng, and
  Zhao}]{chen-etal-2015-event}
Yubo Chen, Liheng Xu, Kang Liu, Daojian Zeng, and Jun Zhao. 2015.
\newblock \href {https://doi.org/10.3115/v1/P15-1017} {Event extraction via
  dynamic multi-pooling convolutional neural networks}.
\newblock In \emph{Proceedings of the 53rd Annual Meeting of the Association
  for Computational Linguistics and the 7th International Joint Conference on
  Natural Language Processing (Volume 1: Long Papers)}, pages 167--176,
  Beijing, China. Association for Computational Linguistics.

\bibitem[{Chen et~al.(2018)Chen, Yang, Liu, Zhao, and
  Jia}]{chen-etal-2018-collective}
Yubo Chen, Hang Yang, Kang Liu, Jun Zhao, and Yantao Jia. 2018.
\newblock \href {https://doi.org/10.18653/v1/D18-1158} {Collective event
  detection via a hierarchical and bias tagging networks with gated multi-level
  attention mechanisms}.
\newblock In \emph{Proceedings of the 2018 Conference on Empirical Methods in
  Natural Language Processing}, pages 1267--1276, Brussels, Belgium.
  Association for Computational Linguistics.

\bibitem[{Chen et~al.(2020)Chen, Chen, Ebner, White, and
  Van~Durme}]{chen-etal-2020-reading}
Yunmo Chen, Tongfei Chen, Seth Ebner, Aaron~Steven White, and Benjamin
  Van~Durme. 2020.
\newblock \href {https://doi.org/10.18653/v1/2020.spnlp-1.9} {Reading the
  manual: Event extraction as definition comprehension}.
\newblock In \emph{Proceedings of the Fourth Workshop on Structured Prediction
  for NLP}, pages 74--83, Online. Association for Computational Linguistics.

\bibitem[{Cui et~al.(2020)Cui, Yu, Liu, Zhang, Wang, and
  Shi}]{cui-etal-2020-edge}
Shiyao Cui, Bowen Yu, Tingwen Liu, Zhenyu Zhang, Xuebin Wang, and Jinqiao Shi.
  2020.
\newblock \href {https://doi.org/10.18653/v1/2020.findings-emnlp.211}
  {Edge-enhanced graph convolution networks for event detection with syntactic
  relation}.
\newblock In \emph{Findings of the Association for Computational Linguistics:
  EMNLP 2020}, pages 2329--2339, Online. Association for Computational
  Linguistics.

\bibitem[{Devlin et~al.(2019)Devlin, Chang, Lee, and
  Toutanova}]{devlin-etal-2019-bert}
Jacob Devlin, Ming-Wei Chang, Kenton Lee, and Kristina Toutanova. 2019.
\newblock \href {https://doi.org/10.18653/v1/N19-1423} {{BERT}: Pre-training of
  deep bidirectional transformers for language understanding}.
\newblock In \emph{Proceedings of the 2019 Conference of the North {A}merican
  Chapter of the Association for Computational Linguistics: Human Language
  Technologies, Volume 1 (Long and Short Papers)}, pages 4171--4186,
  Minneapolis, Minnesota. Association for Computational Linguistics.

\bibitem[{Doddington et~al.(2004)Doddington, Mitchell, Przybocki, Ramshaw,
  Strassel, and Weischedel}]{doddington-etal-2004-automatic}
George Doddington, Alexis Mitchell, Mark Przybocki, Lance Ramshaw, Stephanie
  Strassel, and Ralph Weischedel. 2004.
\newblock \href {http://www.lrec-conf.org/proceedings/lrec2004/pdf/5.pdf} {The
  automatic content extraction ({ACE}) program {--} tasks, data, and
  evaluation}.
\newblock In \emph{Proceedings of the Fourth International Conference on
  Language Resources and Evaluation ({LREC}{'}04)}, Lisbon, Portugal. European
  Language Resources Association (ELRA).

\bibitem[{Du and Cardie(2020)}]{du-cardie-2020-event}
Xinya Du and Claire Cardie. 2020.
\newblock \href {https://doi.org/10.18653/v1/2020.emnlp-main.49} {Event
  extraction by answering (almost) natural questions}.
\newblock In \emph{Proceedings of the 2020 Conference on Empirical Methods in
  Natural Language Processing (EMNLP)}, pages 671--683, Online. Association for
  Computational Linguistics.

\bibitem[{Etzioni et~al.(2005)Etzioni, Cafarella, Downey, Popescu, Shaked,
  Soderland, Weld, and Yates}]{etzioni2005unsupervised}
Oren Etzioni, Michael Cafarella, Doug Downey, Ana-Maria Popescu, Tal Shaked,
  Stephen Soderland, Daniel~S Weld, and Alexander Yates. 2005.
\newblock Unsupervised named-entity extraction from the web: An experimental
  study.
\newblock \emph{Artificial intelligence}, 165(1):91--134.

\bibitem[{Finkel and Manning(2009)}]{finkel-manning-2009-nested}
Jenny~Rose Finkel and Christopher~D. Manning. 2009.
\newblock \href {https://www.aclweb.org/anthology/D09-1015} {Nested named
  entity recognition}.
\newblock In \emph{Proceedings of the 2009 Conference on Empirical Methods in
  Natural Language Processing}, pages 141--150, Singapore. Association for
  Computational Linguistics.

\bibitem[{Huang et~al.(2015)Huang, Xu, and Yu}]{huang2015bidirectional}
Zhiheng Huang, Wei Xu, and Kai Yu. 2015.
\newblock Bidirectional lstm-crf models for sequence tagging.
\newblock \emph{arXiv preprint arXiv:1508.01991}.

\bibitem[{Katiyar and Cardie(2018)}]{katiyar-cardie-2018-nested}
Arzoo Katiyar and Claire Cardie. 2018.
\newblock \href {https://doi.org/10.18653/v1/N18-1079} {Nested named entity
  recognition revisited}.
\newblock In \emph{Proceedings of the 2018 Conference of the North {A}merican
  Chapter of the Association for Computational Linguistics: Human Language
  Technologies, Volume 1 (Long Papers)}, pages 861--871, New Orleans,
  Louisiana. Association for Computational Linguistics.

\bibitem[{Kingma and Ba(2015)}]{DBLP:journals/corr/KingmaB14}
Diederik~P. Kingma and Jimmy Ba. 2015.
\newblock \href {http://arxiv.org/abs/1412.6980} {Adam: A method for stochastic
  optimization}.
\newblock In \emph{ICLR (Poster)}.

\bibitem[{Lafferty et~al.(2001)Lafferty, McCallum, and
  Pereira}]{lafferty2001conditional}
John Lafferty, Andrew McCallum, and Fernando~CN Pereira. 2001.
\newblock Conditional random fields: Probabilistic models for segmenting and
  labeling sequence data.

\bibitem[{Levow(2006)}]{levow-2006-third}
Gina-Anne Levow. 2006.
\newblock \href {https://www.aclweb.org/anthology/W06-0115} {The third
  international {C}hinese language processing bakeoff: Word segmentation and
  named entity recognition}.
\newblock In \emph{Proceedings of the Fifth {SIGHAN} Workshop on {C}hinese
  Language Processing}, pages 108--117, Sydney, Australia. Association for
  Computational Linguistics.

\bibitem[{Levy et~al.(2017)Levy, Seo, Choi, and
  Zettlemoyer}]{levy-etal-2017-zero}
Omer Levy, Minjoon Seo, Eunsol Choi, and Luke Zettlemoyer. 2017.
\newblock \href {https://doi.org/10.18653/v1/K17-1034} {Zero-shot relation
  extraction via reading comprehension}.
\newblock In \emph{Proceedings of the 21st Conference on Computational Natural
  Language Learning ({C}o{NLL} 2017)}, pages 333--342, Vancouver, Canada.
  Association for Computational Linguistics.

\bibitem[{Li et~al.(2020{\natexlab{a}})Li, Peng, Chen, Wang, Pan, Lyu, and
  Zhu}]{li-etal-2020-event}
Fayuan Li, Weihua Peng, Yuguang Chen, Quan Wang, Lu~Pan, Yajuan Lyu, and Yong
  Zhu. 2020{\natexlab{a}}.
\newblock \href {https://doi.org/10.18653/v1/2020.findings-emnlp.73} {Event
  extraction as multi-turn question answering}.
\newblock In \emph{Findings of the Association for Computational Linguistics:
  EMNLP 2020}, pages 829--838, Online. Association for Computational
  Linguistics.

\bibitem[{Li et~al.(2020{\natexlab{b}})Li, Feng, Meng, Han, Wu, and
  Li}]{li-etal-2020-unified}
Xiaoya Li, Jingrong Feng, Yuxian Meng, Qinghong Han, Fei Wu, and Jiwei Li.
  2020{\natexlab{b}}.
\newblock \href {https://doi.org/10.18653/v1/2020.acl-main.519} {A unified
  {MRC} framework for named entity recognition}.
\newblock In \emph{Proceedings of the 58th Annual Meeting of the Association
  for Computational Linguistics}, pages 5849--5859, Online. Association for
  Computational Linguistics.

\bibitem[{Li et~al.(2019)Li, Yin, Sun, Li, Yuan, Chai, Zhou, and
  Li}]{li-etal-2019-entity}
Xiaoya Li, Fan Yin, Zijun Sun, Xiayu Li, Arianna Yuan, Duo Chai, Mingxin Zhou,
  and Jiwei Li. 2019.
\newblock \href {https://doi.org/10.18653/v1/P19-1129} {Entity-relation
  extraction as multi-turn question answering}.
\newblock In \emph{Proceedings of the 57th Annual Meeting of the Association
  for Computational Linguistics}, pages 1340--1350, Florence, Italy.
  Association for Computational Linguistics.

\bibitem[{Lin et~al.(2019{\natexlab{a}})Lin, Lu, Han, and
  Sun}]{lin-etal-2019-cost}
Hongyu Lin, Yaojie Lu, Xianpei Han, and Le~Sun. 2019{\natexlab{a}}.
\newblock \href {https://doi.org/10.18653/v1/P19-1521} {Cost-sensitive
  regularization for label confusion-aware event detection}.
\newblock In \emph{Proceedings of the 57th Annual Meeting of the Association
  for Computational Linguistics}, pages 5278--5283, Florence, Italy.
  Association for Computational Linguistics.

\bibitem[{Lin et~al.(2019{\natexlab{b}})Lin, Lu, Han, and
  Sun}]{lin-etal-2019-sequence}
Hongyu Lin, Yaojie Lu, Xianpei Han, and Le~Sun. 2019{\natexlab{b}}.
\newblock \href {https://doi.org/10.18653/v1/P19-1511} {Sequence-to-nuggets:
  Nested entity mention detection via anchor-region networks}.
\newblock In \emph{Proceedings of the 57th Annual Meeting of the Association
  for Computational Linguistics}, pages 5182--5192, Florence, Italy.
  Association for Computational Linguistics.

\bibitem[{Liu et~al.(2020)Liu, Chen, Liu, Bi, and Liu}]{liu-etal-2020-event}
Jian Liu, Yubo Chen, Kang Liu, Wei Bi, and Xiaojiang Liu. 2020.
\newblock \href {https://doi.org/10.18653/v1/2020.emnlp-main.128} {Event
  extraction as machine reading comprehension}.
\newblock In \emph{Proceedings of the 2020 Conference on Empirical Methods in
  Natural Language Processing (EMNLP)}, pages 1641--1651, Online. Association
  for Computational Linguistics.

\bibitem[{Liu et~al.(2016)Liu, Chen, He, Liu, and
  Zhao}]{liu-etal-2016-leveraging}
Shulin Liu, Yubo Chen, Shizhu He, Kang Liu, and Jun Zhao. 2016.
\newblock \href {https://doi.org/10.18653/v1/P16-1201} {Leveraging {F}rame{N}et
  to improve automatic event detection}.
\newblock In \emph{Proceedings of the 54th Annual Meeting of the Association
  for Computational Linguistics (Volume 1: Long Papers)}, pages 2134--2143,
  Berlin, Germany. Association for Computational Linguistics.

\bibitem[{Liu et~al.(2017)Liu, Chen, Liu, and Zhao}]{liu-etal-2017-exploiting}
Shulin Liu, Yubo Chen, Kang Liu, and Jun Zhao. 2017.
\newblock \href {https://doi.org/10.18653/v1/P17-1164} {Exploiting argument
  information to improve event detection via supervised attention mechanisms}.
\newblock In \emph{Proceedings of the 55th Annual Meeting of the Association
  for Computational Linguistics (Volume 1: Long Papers)}, pages 1789--1798,
  Vancouver, Canada. Association for Computational Linguistics.

\bibitem[{Liu et~al.(2018{\natexlab{a}})Liu, Luo, and Huang}]{liu2018jointly}
Xiao Liu, Zhunchen Luo, and Heyan Huang. 2018{\natexlab{a}}.
\newblock Jointly multiple events extraction via attention-based graph
  information aggregation.
\newblock \emph{arXiv preprint arXiv:1809.09078}.

\bibitem[{Liu et~al.(2018{\natexlab{b}})Liu, Luo, and
  Huang}]{liu-etal-2018-jointly}
Xiao Liu, Zhunchen Luo, and Heyan Huang. 2018{\natexlab{b}}.
\newblock \href {https://doi.org/10.18653/v1/D18-1156} {Jointly multiple events
  extraction via attention-based graph information aggregation}.
\newblock In \emph{Proceedings of the 2018 Conference on Empirical Methods in
  Natural Language Processing}, pages 1247--1256, Brussels, Belgium.
  Association for Computational Linguistics.

\bibitem[{Luan et~al.(2019)Luan, Wadden, He, Shah, Ostendorf, and
  Hajishirzi}]{luan-etal-2019-general}
Yi~Luan, Dave Wadden, Luheng He, Amy Shah, Mari Ostendorf, and Hannaneh
  Hajishirzi. 2019.
\newblock \href {https://doi.org/10.18653/v1/N19-1308} {A general framework for
  information extraction using dynamic span graphs}.
\newblock In \emph{Proceedings of the 2019 Conference of the North {A}merican
  Chapter of the Association for Computational Linguistics: Human Language
  Technologies, Volume 1 (Long and Short Papers)}, pages 3036--3046,
  Minneapolis, Minnesota. Association for Computational Linguistics.

\bibitem[{Ma and Hovy(2016)}]{ma-hovy-2016-end}
Xuezhe Ma and Eduard Hovy. 2016.
\newblock \href {https://doi.org/10.18653/v1/P16-1101} {End-to-end sequence
  labeling via bi-directional {LSTM}-{CNN}s-{CRF}}.
\newblock In \emph{Proceedings of the 54th Annual Meeting of the Association
  for Computational Linguistics (Volume 1: Long Papers)}, pages 1064--1074,
  Berlin, Germany. Association for Computational Linguistics.

\bibitem[{Meng et~al.(2019)Meng, Wu, Wang, Li, Nie, Yin, Li, Han, Sun, and
  Li}]{meng2019glyce}
Yuxian Meng, Wei Wu, Fei Wang, Xiaoya Li, Ping Nie, Fan Yin, Muyu Li, Qinghong
  Han, Xiaofei Sun, and Jiwei Li. 2019.
\newblock Glyce: Glyph-vectors for chinese character representations.
\newblock In \emph{Advances in Neural Information Processing Systems}, pages
  2746--2757.

\bibitem[{Nguyen et~al.(2016)Nguyen, Cho, and
  Grishman}]{nguyen-etal-2016-joint}
Thien~Huu Nguyen, Kyunghyun Cho, and Ralph Grishman. 2016.
\newblock \href {https://doi.org/10.18653/v1/N16-1034} {Joint event extraction
  via recurrent neural networks}.
\newblock In \emph{Proceedings of the 2016 Conference of the North {A}merican
  Chapter of the Association for Computational Linguistics: Human Language
  Technologies}, pages 300--309, San Diego, California. Association for
  Computational Linguistics.

\bibitem[{Pennington et~al.(2014)Pennington, Socher, and
  Manning}]{pennington2014glove}
Jeffrey Pennington, Richard Socher, and Christopher~D Manning. 2014.
\newblock Glove: Global vectors for word representation.
\newblock In \emph{Proceedings of the 2014 conference on empirical methods in
  natural language processing (EMNLP)}, pages 1532--1543.

\bibitem[{Peters et~al.(2018)Peters, Neumann, Iyyer, Gardner, Clark, Lee, and
  Zettlemoyer}]{peters-etal-2018-deep}
Matthew Peters, Mark Neumann, Mohit Iyyer, Matt Gardner, Christopher Clark,
  Kenton Lee, and Luke Zettlemoyer. 2018.
\newblock \href {https://doi.org/10.18653/v1/N18-1202} {Deep contextualized
  word representations}.
\newblock In \emph{Proceedings of the 2018 Conference of the North {A}merican
  Chapter of the Association for Computational Linguistics: Human Language
  Technologies, Volume 1 (Long Papers)}, pages 2227--2237, New Orleans,
  Louisiana. Association for Computational Linguistics.

\bibitem[{Pradhan et~al.(2011)Pradhan, Ramshaw, Marcus, Palmer, Weischedel, and
  Xue}]{pradhan-etal-2011-conll}
Sameer Pradhan, Lance Ramshaw, Mitchell Marcus, Martha Palmer, Ralph
  Weischedel, and Nianwen Xue. 2011.
\newblock \href {https://www.aclweb.org/anthology/W11-1901} {{C}o{NLL}-2011
  shared task: Modeling unrestricted coreference in {O}nto{N}otes}.
\newblock In \emph{Proceedings of the Fifteenth Conference on Computational
  Natural Language Learning: Shared Task}, pages 1--27, Portland, Oregon, USA.
  Association for Computational Linguistics.

\bibitem[{Sekine and Nobata(2004)}]{sekine2004definition}
Satoshi Sekine and Chikashi Nobata. 2004.
\newblock Definition, dictionaries and tagger for extended named entity
  hierarchy.
\newblock In \emph{LREC}. Lisbon, Portugal.

\bibitem[{Shibuya and Hovy(2020)}]{shibuya-hovy-2020-nested}
Takashi Shibuya and Eduard Hovy. 2020.
\newblock \href {https://doi.org/10.1162/tacl_a_00334} {Nested named entity
  recognition via second-best sequence learning and decoding}.
\newblock \emph{Transactions of the Association for Computational Linguistics},
  8:605--620.

\bibitem[{Strakov{\'a} et~al.(2019)Strakov{\'a}, Straka, and
  Hajic}]{strakova-etal-2019-neural}
Jana Strakov{\'a}, Milan Straka, and Jan Hajic. 2019.
\newblock \href {https://doi.org/10.18653/v1/P19-1527} {Neural architectures
  for nested {NER} through linearization}.
\newblock In \emph{Proceedings of the 57th Annual Meeting of the Association
  for Computational Linguistics}, pages 5326--5331, Florence, Italy.
  Association for Computational Linguistics.

\bibitem[{Strubell et~al.(2017)Strubell, Verga, Belanger, and
  McCallum}]{strubell2017fast}
Emma Strubell, Patrick Verga, David Belanger, and Andrew McCallum. 2017.
\newblock Fast and accurate entity recognition with iterated dilated
  convolutions.
\newblock \emph{arXiv preprint arXiv:1702.02098}.

\bibitem[{Sutton et~al.(2007)Sutton, McCallum, and
  Rohanimanesh}]{sutton2007dynamic}
Charles Sutton, Andrew McCallum, and Khashayar Rohanimanesh. 2007.
\newblock Dynamic conditional random fields: Factorized probabilistic models
  for labeling and segmenting sequence data.
\newblock \emph{Journal of Machine Learning Research}, 8(Mar):693--723.

\bibitem[{Tong et~al.(2020)Tong, Xu, Wang, Cao, Hou, Li, and
  Xie}]{tong-etal-2020-improving}
Meihan Tong, Bin Xu, Shuai Wang, Yixin Cao, Lei Hou, Juanzi Li, and Jun Xie.
  2020.
\newblock \href {https://doi.org/10.18653/v1/2020.acl-main.522} {Improving
  event detection via open-domain trigger knowledge}.
\newblock In \emph{Proceedings of the 58th Annual Meeting of the Association
  for Computational Linguistics}, pages 5887--5897, Online. Association for
  Computational Linguistics.

\bibitem[{Vaswani et~al.(2017)Vaswani, Shazeer, Parmar, Uszkoreit, Jones,
  Gomez, Kaiser, and Polosukhin}]{vaswani2017attention}
Ashish Vaswani, Noam Shazeer, Niki Parmar, Jakob Uszkoreit, Llion Jones,
  Aidan~N Gomez, Lukasz Kaiser, and Illia Polosukhin. 2017.
\newblock Attention is all you need.
\newblock In \emph{NIPS}.

\bibitem[{Wang and Lu(2018)}]{wang-lu-2018-neural}
Bailin Wang and Wei Lu. 2018.
\newblock \href {https://doi.org/10.18653/v1/D18-1019} {Neural segmental
  hypergraphs for overlapping mention recognition}.
\newblock In \emph{Proceedings of the 2018 Conference on Empirical Methods in
  Natural Language Processing}, pages 204--214, Brussels, Belgium. Association
  for Computational Linguistics.

\bibitem[{Wang et~al.(2018)Wang, Qu, Chen, Shen, Zhang, Zhang, Gao, Gu, Chen,
  and Yu}]{wang-etal-2018-label}
Zhenghui Wang, Yanru Qu, Liheng Chen, Jian Shen, Weinan Zhang, Shaodian Zhang,
  Yimei Gao, Gen Gu, Ken Chen, and Yong Yu. 2018.
\newblock \href {https://doi.org/10.18653/v1/N18-1001} {Label-aware double
  transfer learning for cross-specialty medical named entity recognition}.
\newblock In \emph{Proceedings of the 2018 Conference of the North {A}merican
  Chapter of the Association for Computational Linguistics: Human Language
  Technologies, Volume 1 (Long Papers)}, pages 1--15, New Orleans, Louisiana.
  Association for Computational Linguistics.

\bibitem[{Wei et~al.(2020)Wei, Su, Wang, Tian, and Chang}]{wei-etal-2020-novel}
Zhepei Wei, Jianlin Su, Yue Wang, Yuan Tian, and Yi~Chang. 2020.
\newblock \href {https://doi.org/10.18653/v1/2020.acl-main.136} {A novel
  cascade binary tagging framework for relational triple extraction}.
\newblock In \emph{Proceedings of the 58th Annual Meeting of the Association
  for Computational Linguistics}, pages 1476--1488, Online. Association for
  Computational Linguistics.

\bibitem[{Yang et~al.(2019{\natexlab{a}})Yang, Feng, Qiao, Kan, and
  Li}]{yang-etal-2019-exploring}
Sen Yang, Dawei Feng, Linbo Qiao, Zhigang Kan, and Dongsheng Li.
  2019{\natexlab{a}}.
\newblock \href {https://doi.org/10.18653/v1/P19-1522} {Exploring pre-trained
  language models for event extraction and generation}.
\newblock In \emph{Proceedings of the 57th Annual Meeting of the Association
  for Computational Linguistics}, pages 5284--5294, Florence, Italy.
  Association for Computational Linguistics.

\bibitem[{Yang et~al.(2019{\natexlab{b}})Yang, Feng, Qiao, Kan, and
  Li}]{yang-etal-2019-exploring-pre}
Sen Yang, Dawei Feng, Linbo Qiao, Zhigang Kan, and Dongsheng Li.
  2019{\natexlab{b}}.
\newblock \href {https://doi.org/10.18653/v1/P19-1522} {Exploring pre-trained
  language models for event extraction and generation}.
\newblock In \emph{Proceedings of the 57th Annual Meeting of the Association
  for Computational Linguistics}, pages 5284--5294, Florence, Italy.
  Association for Computational Linguistics.

\bibitem[{Yu et~al.(2020)Yu, Bohnet, and Poesio}]{yu-etal-2020-named}
Juntao Yu, Bernd Bohnet, and Massimo Poesio. 2020.
\newblock \href {https://doi.org/10.18653/v1/2020.acl-main.577} {Named entity
  recognition as dependency parsing}.
\newblock In \emph{Proceedings of the 58th Annual Meeting of the Association
  for Computational Linguistics}, pages 6470--6476, Online. Association for
  Computational Linguistics.

\bibitem[{Zhang and Yang(2018)}]{zhang-yang-2018-chinese}
Yue Zhang and Jie Yang. 2018.
\newblock \href {https://doi.org/10.18653/v1/P18-1144} {{C}hinese {NER} using
  lattice {LSTM}}.
\newblock In \emph{Proceedings of the 56th Annual Meeting of the Association
  for Computational Linguistics (Volume 1: Long Papers)}, pages 1554--1564,
  Melbourne, Australia. Association for Computational Linguistics.

\end{thebibliography}

\clearpage
\appendix
\section{Appendix}
\subsection{Details of the Heuristic Match Principle}
\label{appendix:Heuristic Match Principle}

\renewcommand{\algorithmicrequire}{\textbf{In:}}
\renewcommand{\algorithmicensure}{\textbf{Out:}}

\renewcommand{\algorithmicrequire}{\textbf{In:}}
\renewcommand{\algorithmicensure}{\textbf{Out:}}
\begin{algorithm}[!h]
    \caption{span determination~\citep{yang-etal-2019-exploring}}
    \begin{algorithmic}[1] 
        \Require $\text{start}_{c}$, $\text{end}_{c}$, and sequence length $l$.
        \Ensure Result list $L$ that each item is a span plays $c^{th}$ category. \\
        \textbf{Initiate: $a_s \leftarrow-1 $, $a_e \leftarrow-1 $}
        \For{$i \leftarrow 0$ to $l$}
        \If{In state 1 \textbf{and} $\text{start}_{x_i}^{c}>0.5$}
        \State $a_s \leftarrow i$ and change to state 2
        \EndIf
        \If{In state 2}
        \If{the $\text{start}_{x_i}^{c} > 0.5$}
        \State $a_{s} \leftarrow i$ \textbf{if} $\text{start}_{x_i}^{c}> \text{start}_{a_{s}}^{c}$
        \EndIf
        \If{$\text{end}_{x_i}^{c} > 0.5$}
        \State $a_e \leftarrow i$ and change to state 3
        \EndIf
        \EndIf
        \If{In state 3}
        \If{$\text{end}_{x_i}^{c}>0.5$}
        \State $a_e\leftarrow i$ \textbf{if} $\text{end}_{x_i}^{c}> \text{end}_{a_{e}}^{c}$
        \EndIf
        \If{$\text{start}_{x_i}^{c}>0.5$}
        \State Append $[a_s,a_e]$ to $L$
        \State $a_s\leftarrow-1$, $a_e\leftarrow i$ and change to state 2
        \EndIf
        \EndIf
        \EndFor
    \end{algorithmic}
    \label{algorithm 1}
\end{algorithm}

Algorithm \ref{algorithm 1} contains a finite state machine, which changes from one state to another in response to $\text{start}^{c}$, $\text{end}^{c}$. There are three states totally: 1) Neither start nor end has been detected; 2) Only a start has been detected; 3) A start as well as an end have been detected. Specially, the state changes according to the following rules: State 1 changes to State 2 when the current token is a start; State 2 changes to State 3 when the current token is an end; State 3 changes to State 2 when the current token is a new start. Notably, if there has been a start and another start arises, we will choose the one with higher probability, and the same for end.

\subsection{Effect of Span Decoding Strategy}
\label{section:span_decoding_strategy}

Table \ref{Tab. span_decoding_strategy} shows the effect of the different span decoding strategies. All of them use the BERT encoder as backbone. The differences are (1) Strategy \textbf{A} treats span decoding as a multi-label classification problem with $2\times |C|$ binary classifiers, which aims to predict the boundary of a span. This strategy is inspired by the QA task and it is adopted in BERT-span and our LEAR. BERT-span$_{v1}$ employs the heuristic match principle, and BERT-span$_{v2}$ uses the nearest match principle, both mentioned in section \ref{section. span decoding}. (2) The most commonly-used Strategy \textbf{B} treats span decoding as a multi-class classification problem with BIO or BIOS schema, and is adopted in BERT-softmax and BERT-crf. Compared with BERT-softmax, BERT-crf adds a conditional random field (CRF) layer to model the dependencies between predictions, usually yielding better performance but worse efficiency. 

\begin{table}[!tb] \small
\centering
\begin{tabular}{cll}
\toprule
\multicolumn{3}{c}{\textbf{OntoNotes 4.0 for NER}}                        \\ \midrule
\multicolumn{1}{l}{\textbf{Strategy}} & \textbf{Method}  & \textbf{F1} \\
\midrule
\multirow{2}{*}{\textbf{A}} & BERT-span$_{v1}$  & 82.65 \\
& BERT-span$_{v2}$ & 82.14 \\
\midrule
\multirow{2}{*}{\textbf{B}} & BERT-crf & 81.65 \\
\cmidrule(){2-3}
& BERT-softmax  & 81.30 \\
\bottomrule
\end{tabular}
\caption{Results with different span decoding strategies. BERT-span$_{v1}$ is the LEAR$_{\text{w/o}}$ mentioned above.}
\label{Tab. span_decoding_strategy}
\end{table}

The results show that: (1) The strategy used by LEAR has better performance than the traditional way. The reason might be that, the span decoding strategy in our approach is start/end position matching, which only needs to predict the span's boundary. In contrast, the strategy adopted in previous methods needs to predict both boundary and internal words, which is much harder, especially for a longer span. (2) The comparison between BERT-span$_{v1}$ and BERT-span$_{v2}$ shows that, the heuristic match principle could achieve better results by making the most of information from probability. (3) Besides, there is an extra benefit for Strategy \textbf{A}. It naturally tackles the nested span issue, which means that candidate span overlaps with different categories.

\subsection{Details of Hyper-Parameters Settings}
\label{appendix:hyper-parameters}
All hyper-parameters of our model are listed in Table~\ref{tab:hyper-paras} in detail.
\begin{table*}[!tp]\small
\centering
\begin{tabular}{@{}cccccccc@{}}
\toprule
\multirow{2}{*}{\textbf{}} & \multirow{2}{*}{\textbf{random seed}} & \multirow{2}{*}{\textbf{max\_seq\_len}} & \multirow{2}{*}{\textbf{batch size}} & \multirow{2}{*}{\textbf{epoch}} & \multirow{2}{*}{\textbf{dropout rate}} & \multicolumn{2}{c}{\textbf{learning rate}}   \\ \cmidrule(l){7-8} 
                          &                                &                                         &                                      &                                 &                                        & \textbf{encoder layer} & \textbf{task layer} \\ \cmidrule(r){1-8}
ACE 2005 (ED)              & 1                              & 256                            & 32                          & 30                              & 0.1                                    & 1e-5                   & 2e-4                \\
ACE 2005 (NER)             & 42                             & 128                                     & 32                                   & 20                              & 0.1                                    & 3e-5                   & 6e-5                \\
ACE 2004                   & 42                             & 128                                     & 32                                   & 30                              & 0.1                                    & 3e-5                   & 3e-4                \\
OntoNotes 4.0              & 42                             & 128                                     & 32                                   & 5                               & 0.1                                    & 8e-6                   & 8e-5                \\
MSRA                       & 42                             & 128                                     & 32                                   & 20                              & 0.1                                    & 3e-5                   & 6e-5                \\ \bottomrule
\end{tabular}
\caption{Hyper-parameter settings for each experiment. }
\label{tab:hyper-paras}
\end{table*}

\subsection{Statistics of the datasets used in the experiments}\label{sec:data_statistics}
\begin{table*}[!t]
\centering
\resizebox{\textwidth}{!}{%
\begin{tabular}{@{}cl|ccc|ccc|ccc|ccc|ccc@{}}
\toprule
\multicolumn{2}{c}{\multirow{2}{*}{}}                                 & \multicolumn{3}{c}{\textbf{ACE 2005(ED)}}     & \multicolumn{3}{c}{\textbf{ACE2005(NER)}}     & \multicolumn{3}{c}{\textbf{ACE2004}}          & \multicolumn{3}{c}{\textbf{OntoNote 4.0(Chinese)}}   & \multicolumn{3}{c}{\textbf{MSRA}}                                  \\ \cmidrule(){3-17} 
\multicolumn{2}{c}{}                                                  & \textbf{train} & \textbf{dev} & \textbf{test} & \textbf{train} & \textbf{dev} & \textbf{test} & \textbf{train} & \textbf{dev} & \textbf{test} & \textbf{train} & \textbf{dev} & \textbf{test}        & \textbf{train}       & \textbf{dev}         & \textbf{test}        \\ \cmidrule(){1-17}
\multicolumn{2}{c}{\textbf{sentences}}                                & 13919          & 880          & 810           & 7294           & 971          & 1057          & 6200           & 745          & 812           & 15650          & 4301         & 4346                 & 41729                & 4637                 & 4366                 \\ \cmidrule(){1-17}
\multirow{2}{*}{\textbf{spans}} & \textbf{\# total}                   & 4496           & 279          & 574           & 24703          & 3218         & 3029          & 22201          & 2514         & 3035          & 13367          & 6950         & 7684                 & 70446                & 4157                 & 6181                 \\
                                & \multirow{1}{*}{\textbf{\# nested}} & -              & -            & -             & 5052           & 598          & 638           & 5416           & 623          & 779           & -              & -            & -                    & -                    & -                    & -                    \\
 &  & & &  & (20.45\%) & (18.58\%) & (21.06\%) & (24.40\%) & (24.78\%) & (25.67\%) & &  &  &  &  & \\\bottomrule
\end{tabular}%
}
\caption{Statistics of the datasets used in the experiments. Spans are considered nested only if they are overlapped or nested in the different category.}
\label{tab:statistics}
\end{table*}

Table~\ref{tab:statistics} shows the statistics of the datasets used in the experiments. For ACE2005~(ED), we refer to the previous work\footnote{\url{https://github.com/thunlp/HMEAE}} to process raw data, which follows standard data splitting strategy. NER datasets we used are provided in the previous SOTA work\footnote{\url{https://github.com/ShannonAI/mrc-for-flat-nested-ner}}.

\end{document}